\documentclass{article}

\usepackage{PRIMEarxiv}

\usepackage[utf8]{inputenc} 
\usepackage[T1]{fontenc}    
\usepackage{hyperref}       
\usepackage{url}            
\usepackage{booktabs}       
\usepackage{amsfonts}       
\usepackage{nicefrac}       
\usepackage{microtype}      
\usepackage{lipsum}
\usepackage{fancyhdr}       
\usepackage{graphicx}       
\usepackage{xcolor}
\graphicspath{{media/}}     
\usepackage{amsmath}
\usepackage{hyperref}
\usepackage{natbib}
\usepackage{float}
\usepackage{subcaption}
\setcitestyle{authoryear,open={(},close={)}} 

\newcommand{\longname}{Parallel Attention and Feed-Forward Net Design }

\newcommand{\name}{PAF }
\newcommand{\namens}{PAF}
\newcommand{\longsaf}{Series Attention and Feed-Forward Net Design }

\newcommand{\saf}{SAF }
\newcommand{\safns}{SAF}
\newcommand\mc{\mathcal}
\newcommand\mbb{\mathbb}
\newcommand\mb{\mathbf}

\newcommand\tr{\textrm}

\title{
Investigating the Role of Feed-Forward Networks \\ in Transformers using Parallel Attention and Feed-Forward Net Design
}

\author{
  Shashank Sonkar, Richard G. Baraniuk \\
  Rice University \\
  Houston\\
  \texttt{\{ss164, richb\}@rice.edu} \\
}

\begin{document}
\maketitle

\begin{abstract}
This paper investigates the key role of Feed-Forward Networks (FFNs) in transformer models by utilizing the Parallel Attention and Feed-Forward Net Design (PAF) architecture, and comparing it to their Series Attention and Feed-Forward Net Design (SAF) counterparts. Central to the effectiveness of PAF are two main assumptions regarding the FFN block and the attention block within a layer: 1) the primary function of the FFN block is to maintain isotropy among token embeddings and prevent their degeneration, and 2) the residual norm computed in the attention block is substantially smaller than the input token embedding norm. To empirically validate these assumptions, we train PAF variants of two large language models (RoBERTa-large and bert-large-uncased). Our results demonstrate that both assumptions hold true in the PAF design. This study contributes to a deeper understanding of the roles and interactions between FFNs and self-attention mechanisms in transformer architectures.

\end{abstract}

\section{Introduction}
In recent years, the field of natural language processing (NLP) has witnessed substantial advancements due to the emergence of deep learning and the availability of vast amounts of data. One of the most significant breakthroughs is the transformer model, which has achieved state-of-the-art results in various NLP tasks, such as language translation \citep{mtcite1,raganato2018analysis,liu2020very}, text classification \citep{tccite1,chang2019x,sun2019fine,chang2020taming}, and question answering \citep{lukovnikov2019pretrained,t5,cao2020deformer}. The transformer architecture, introduced by \cite{attention}, in the seminal paper `Attention is All You Need', has revolutionized the NLP landscape and greatly enhanced the performance of numerous applications.

Transformer architecture consists of several layers, each of which includes two main components: a self-attention block and a feed-forward neural network (FFN).
The self-attention mechanism computes the attention weights between all pairs of positions in the input sequence and uses them to compute a weighted sum of the relevant information. 
The feed-forward network processes the output of the self-attention mechanism to generate a new representation for each position in the sequence.
Both components use residual connections \citep{resnet} and layer normalization \citep{batchnormandlayernorm} to improve performance and stability.
Despite the significant success of transformer models, the precise roles of their components, particularly the Feed-Forward Network (FFN) blocks, are not yet fully comprehended. 

\begin{figure*}[t]
    \centering
    \includegraphics[scale=0.7]{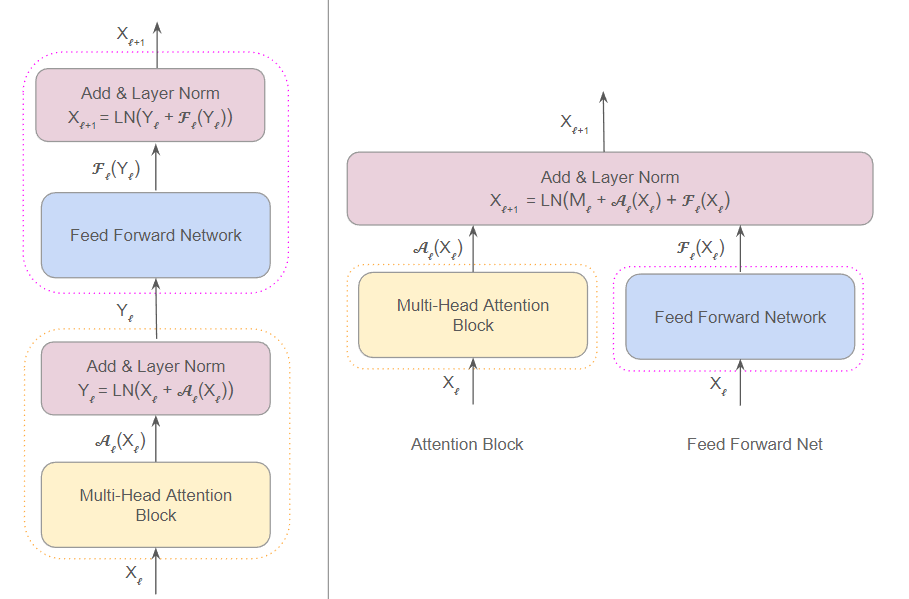}
    \caption{On the left is the standard \longsaf (\safns) for transformers models. On the right is the \longname (\namens) used in transformer models like PaLM \citep{chowdhery2022palm} and Mesh-Transformers \citep{mesh-transformer-jax}.}
    \label{fig:pfa}
\end{figure*}

In this study, we strive to shed light on the functions of these components in transformer architectures by examining the Parallel Attention and Feed-Forward Net Design (PAF), initially proposed in Mesh-Transformers by\cite{mesh-transformer-jax}, subsequently employed by PaLM \citep{chowdhery2022palm}. Contrary to the Series Attention and Feed-Forward Net Design (SAF), PAF facilitates parallelization by having the attention block and FFN block within each layer of the transformer model run concurrently (figure~\ref{fig:pfa}).

In our analysis, we make two critical assumptions based on the PAF architecture: 1) drawing upon the findings from \citep{transformers-token-uniformity,isotropy1}, we posit that the principal function of the FFN block is to prevent the degeneration of token embeddings into a single embedding; and 2) the residual norm computed by the attention block is considerably smaller than the input token embedding norm. To empirically validate these assumptions, we train PAF variants of two prominent language models, RoBERTa-large \citep{roberta} and bert-large-uncased \citep{bert} , and compare their performance to their SAF counterparts on the General Language Understanding (GLUE) benchmark, covering textual entailment, sentiment analysis, and paraphrase detection. Our results reveal the validity of our assumptions on these PAF variants reinforcing our understanding of the FFN's role in maintaining isotropy in token embeddings.

The paper is structured as follows: section \ref{sec:design} outlines the  \name design, section \ref{sec:rat} deep dives into the assumptions and rationale of \name design, and then we conclude in section \ref{sec:con}.

\section{Related work: \longname}
\label{sec:design}
\subsection{\name}
In this section, we first introduce the \name design for parallelization of attention and FFN blocks used in transformer models like PaLM \cite{chowdhery2022palm} and Mesh-Transformers \cite{mesh-transformer-jax}.
\subsubsection{Design changes in \namens:}

At first let's see the computation in standard transformer models which we call the \longsaf (\safns).
Let the input to a standard transformer $\mc{T}$ at layer $l$ be $\mb{X}_l \in \mbb{R}^{n \times d}$.
Let $\mc{T} = \{\mc{A}_i, \mc{F}_i \} $ where $ 0 \leq i \le L$, $L$ is the number of layers, and $\mc{A}, \mc{F}$ are attention and FFN blocks respectively.
Then,
\begin{align}
    \mb{X}_{l+1} &= \tr{LN}\big( \mb{Y}_{l}  + \mc{F}_l(\mb{Y}_{l} ) \big), \; where \label{eq:saf1}\\    
    \mb{Y}_{l} &= \tr{LN}\big( \mb{X}_{l} + \mc{A}_l(\mb{X}_{l}) \big),
    \label{eq:saf2}
\end{align}
where $\tr{LN}$ is layer norm operator. 

\subsubsection*{\name design:} \longname changes the operations of a standard transformer as follows:
\begin{align}
    \mb{X}_{l+1} = \tr{LN}\big( \mb{X}_{l}  +  \mc{A}_l(\mb{X}_{l}) + \mc{F}_l(\mb{X}_{l} ) \big).
    \label{eq:paf}
\end{align}

Note that in the \saf design, the input to the FFN block $\mc{F}_l$ which is $\mb{Y}_{l}$ (equation~\ref{eq:saf1}) relies on the output of the attention block $\mc{A}_l(\mb{X}_{l})$ (equation~\ref{eq:saf2}) thus making the \saf design impossible to parallelize.

\section{Underlying Assumptions of PAF Design}
\label{sec:rat}
\begin{figure}[t]
    \centering
    \begin{subfigure}[b]{0.45\textwidth}
        \centering
        \includegraphics[width=\textwidth]{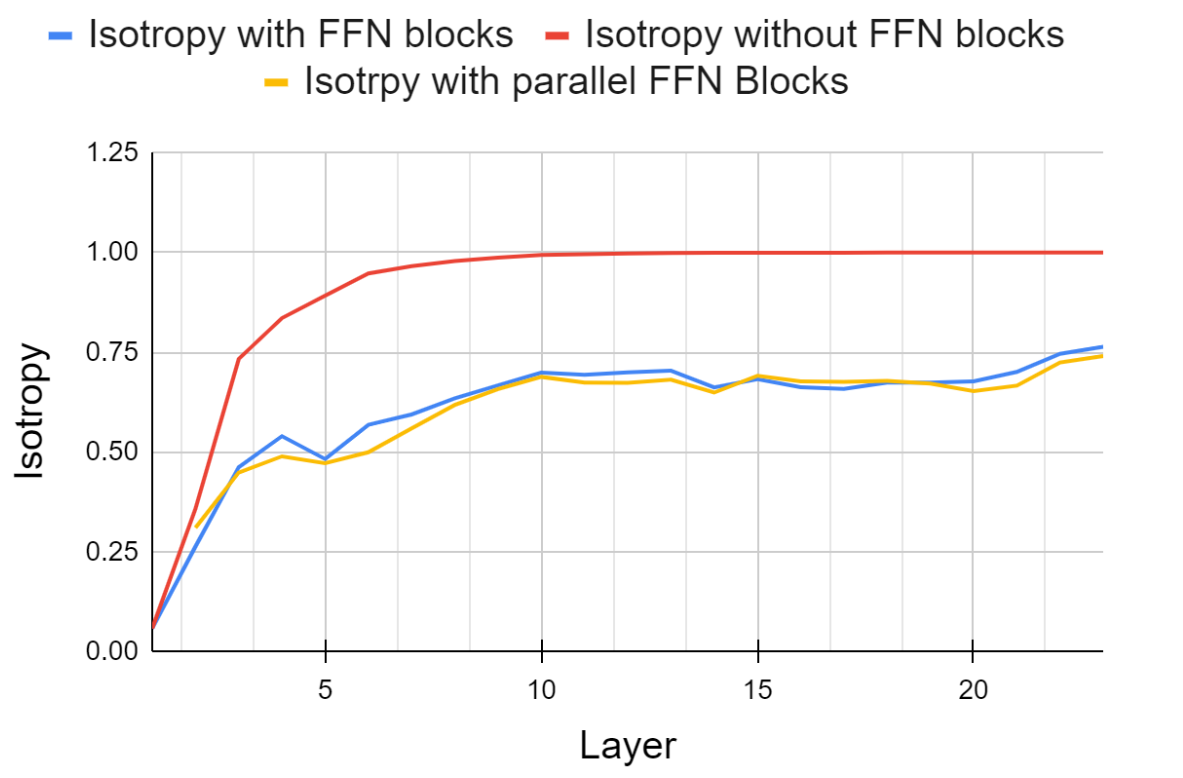}
        \caption{Isotropy measures the closeness of token embeddings at a layer. As can be seen (red) for transformer models (here RoBERTa-large) without FFN blocks, all token embeddings collapse to one single embedding after only a few layers, thus losing any identity information. For both \saf (blue) and \name (yellow) models, FFNs successfully prevent the token embeddings collapse to one single embedding.}
        \label{fig:isosub}
    \end{subfigure}
    \hfill
    \begin{subfigure}[b]{0.5\textwidth}
        \centering
        \includegraphics[width=\textwidth]{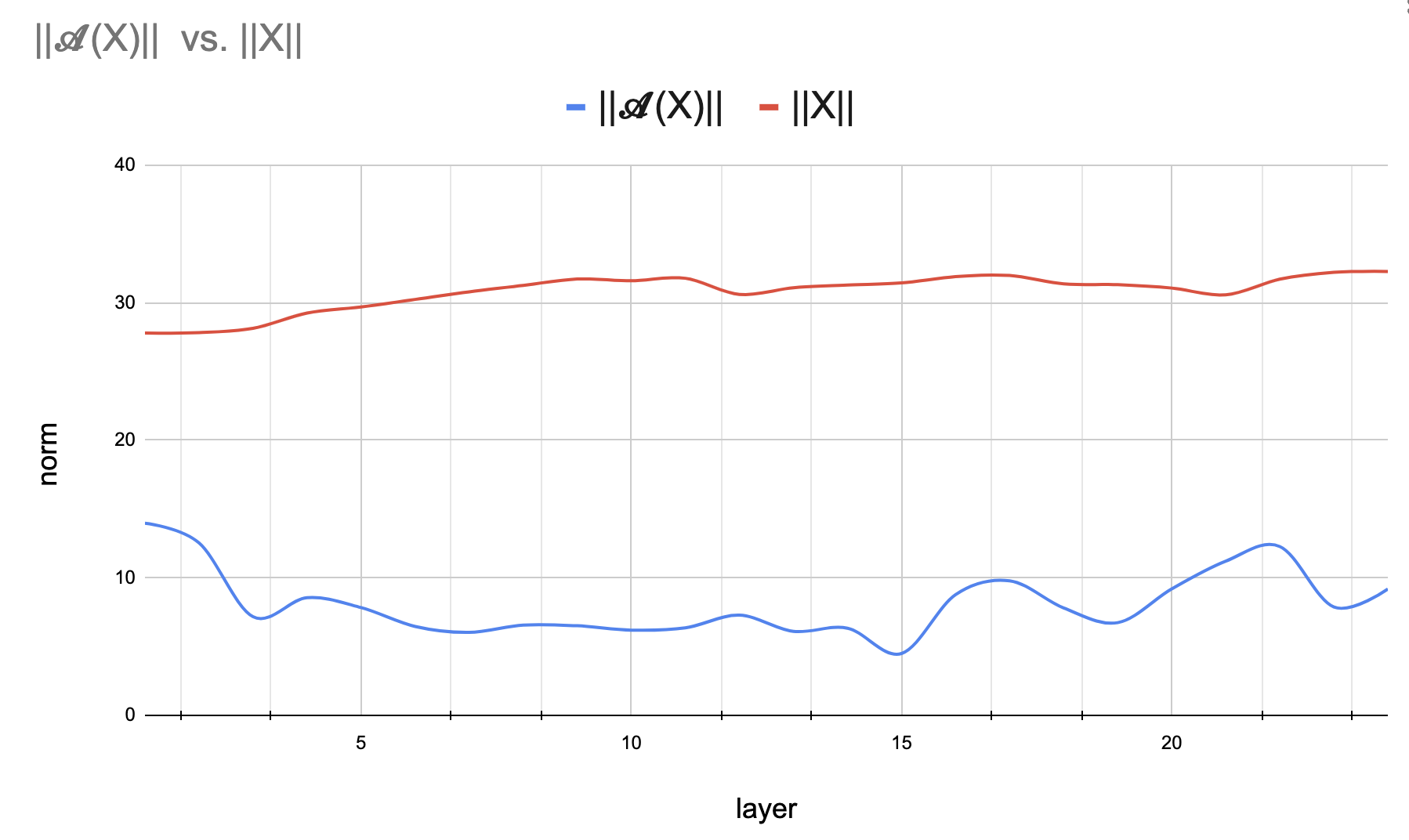}
        \caption{For \name design to work, FFNs need to perform their most important role of preventing token embeddings collapse by spreading out token embeddings (see left figure~\ref{fig:isosub}). Input to the FFN blocks in the \name design is $\mb{X}_{l}$, rather than $\mb{X}_{l} + \mc{A}_l(\mb{X}_{l})$ which is the case in standard \saf design(equation~\ref{eq:paf} vs ~\ref{eq:saf2}). Here we show that $||\mc{A}_l(\mb{X}_{l}||$ is sufficiently small compared to $||\mb{X}_{l}||$ and hence spreading out $\mb{X}_{l}$, rather than $\mb{X}_{l} + \mc{A}_l(\mb{X}_{l})$ also works in practice.}
        \label{fig:resnormsub}
    \end{subfigure}
    \hfill
\end{figure}

In this section, we delve into the reasoning that might explain why the PAF design is as effective as its SAF counterpart. We believe that the PAF design operates on two primary assumptions:

\name makes two assumptions:
\begin{enumerate}
    \item Main function of a FFN block is to maintain isotropy within a layer, i.e., spread out the token embeddings so that the embeddings do not converge to one single embedding, and thereby token embeddings do not lose individual token information.

    \item The norm of the residual computed by a attention block that gets added to input token embedding to the attention block is sufficiently small compared to the norm of the input token embedding.
\end{enumerate}
 
Though the success of \name design itself validates the assumptions, next we provide more evidence to justify these assumptions.

\subsection*{Assumption 1: Role of FFN block in transformers is to prevent degeneration of token embeddings}

\cite{transformers-token-uniformity} show that the token embeddings in transformers models without skip connections and FFN blocks degenerate to a rank-1 matrix doubly exponentially with depth.
The authors present a theoretical argument demonstrating the importance of skip connections in reducing degeneration and suggest that the FFN block can assist in slowing this process.
However, it is important to note that the study does not provide definitive evidence that slowing down degeneration is the most critical or indispensable function of the FFN block.
Further research is necessary to determine the full extent of the role played by the FFN block in transformer models.

In this paper, we make a strong assumption that the main role of an FFN block is to counteract degeneration of token embeddings.
The success/ failure of our experiments will thus validate/ undermine our assumption.
Unlike \cite{transformers-token-uniformity}, we study the degeneration through the lens of isotropy as done by \citet{isotropy1}.

Isotropy measures the average distance between the embedding of each token and the embeddings of the other tokens.
Isotropy $I: \mbb{R}^{n \times d} \to \mbb{R}$ for an embedding matrix $\mb{E} \in \mbb{R}^{n \times d}$ is given by:
\begin{equation}
    I(\mb{E}) = \sum_{0 \leq i < n} \sum_{0 \leq j < n} \frac{E_i^TE_j}{n^2\times||E_i||\times||E_j||}.
\end{equation}

\textbf{Effectiveness of \name to counteract degeneration:}
For a transformer without FFN blocks, isotropy of token embeddings at layer $ I(\mb{X}_l)$ rapidly approaches 1 after few layers of computation as can be seen in figure~\ref{fig:isosub}.
Also, figure~\ref{fig:isosub} shows the effectiveness of \name design to maintain isotropy is at par with \saf design.

\subsection*{Assumption 2: Norm of the attention block's residual is sufficiently the norm of the input token embeddings to the attention block:}

If the main role of FFN blocks is to maintain isotropy by spreading out token embeddings $\mb{Y}_l$ at layer $l$ and \name feeds the input of the attention block $\mb{X}_l$ to the FFN block rather than its output $\mb{Y}_l$ (equations~\eqref{eq:saf2}-~\eqref{eq:paf} and figure~\ref{fig:pfa}), it is imperative to show that $\mb{X}_l$ and $\mb{Y}_l$ are close in the high dimensional space.
In other words, the residual $\mc{A}_l(\mb{X}_{l})$ added to $\mb{X}_l$ by the attention block is small.
If it were not the case, FFN's spreading out $\mb{X}_l$ instead of $\mb{Y}_l$ would not work.
In figure~\ref{fig:resnormsub}, we plot the norm of $\mb{X}_l$ and $\mc{A}_l(\mb{X}_{l})$ for all layers of RoBERTa-large model and find that it is indeed the case.


\begin{table*}[t]
\centering
\resizebox{0.9\textwidth}{!}{
\begin{tabular}{|c|c|c|c|c|c|c|c|}
\hline
\textbf{} & \textbf{MRPC} & \textbf{STS-B} & \textbf{SST-2} & \textbf{QNLI} & \textbf{QQP} & \textbf{MNLI} & \textbf{Avg.} \\ \hline
\textbf{RoBERTa-large} & 90.9 & 92.4 & 96.4 & 94.7 & 92.2 & 90.2 & 92.8 \\ \hline
\textbf{\begin{tabular}[c]{@{}c@{}}RoBERTa-large\\ (w. \namens)\end{tabular}} & 90.5 & 91.0 & 96.2 & 94.3 & 91.7 & 89.3 & 92.2 \\ \hline
\textbf{Bert-Large-Uncased} & 85.0 & 89.2 & 93.5 & 92.2 & 91.4 & 86.6 & 89.6 \\ \hline
\textbf{\begin{tabular}[c]{@{}c@{}}Bert-Large-Uncased\\ (w. \namens)\end{tabular}} & 86.8 & 88.8 & 93.5 & 91.4 & 91.2 & 85.5 & 89.5 \\ \hline
\end{tabular}
}
\caption{This table highlights the effectiveness of the \longname(\namens) variants of RoBERTa-large and Bert-Large-Uncased on the GLUE benchmark. For both models, \name variants perform similarly to the standard \saf equivalents. Note that the gap in RoBERTa is slightly larger than Bert ($0.6\%$ vs $0.1\%$), but the \name variant of RoBERTa has been trained on 10 times less data than the \saf model. For Bert, both the \saf and \name variants use the same size of training data.}
\label{tab:glue}
\end{table*}

\subsection*{Pre-training of \name models}
To fairly compare the both the \saf and \name counterparts to test our assumptions, we pre-trained two large language models RoBERTa-Large \cite{roberta} and Bert-Large-Uncased \cite{bert} on English Wikipedia and BooksCorpus \citep{bookcorpus}.
Both models are 24 layer models and widely used in various NLP applications.
We initialize the parameters for \name models using their \saf variants and follow guidelines for learning rate, optimizer, and loss functions\footnote{\url{https://tinyurl.com/495vfeh9}}.
Each model is trained on four NVIDIA RTX A6000 gpus for a total of 72 hours.

\subsubsection*{Fine-tuning details on the GLUE benchmark}
We tested the effectiveness of the pre-trained \name variants of RoBERTa-Large and Bert-Large-Uncased, we finetune both models on the General Language Understanding Evaluation (GLUE) benchmark \cite{glue}.
GLUE benchmark assesses various NLP tasks that range textual entailment (MNLI, QNLI), paraphrase detection (MRPC, QQP), sentence similarity (STS-B) and sentiment analysis (SST-2).
The GLUE benchmark is a widely recognized evaluation standard for NLP models and provides a comprehensive evaluation of the performance of NLP models.
Each task in GLUE is trained using the recommended\footnote{\url{https://tinyurl.com/26xd6js6}} hyperparameter choices which include learning rate, batch size, warmup steps, and optimizer settings on a single Quadro RTX 8000 GPU for five random seeds.
We exclude the two smallest datasets of the GLUE benchmark - CoLA and RTE because of the high instability and variance in their fine-tuning \citep{unstablebert}.

\subsection*{\name evaluation on GLUE benchmark}
As can be seen in table \ref{tab:glue}, \name variants of both RoBERTa-Large and Bert-Large-Uncased perform nearly identically to their \saf equivalents.
The gap for RoBERTa-Large is slightly less smaller than Bert-Large-Uncased ($0.6\%$ vs $0.1\%$) which can be attributed to eight times smaller size of data used to train the \name variant of RoBERTa-Large.
RoBERTa models were trained on 160GB size dataset, however we only use 20 GB wikipedia and BooksCorpus dataset.

\section{Conclusion}
\label{sec:con}
In summary, this research offers valuable insights into the essential roles and interplay between Feed-Forward Networks (FFNs) and self-attention mechanisms in transformers by examining the Parallel Attention and Feed-Forward Net Design (PAF) architecture. The empirical validation conducted on two well-known language models, RoBERTa-large and bert-large-uncased, indicates that both main assumptions regarding the function of FFN blocks and the residual norm of the attention block hold true in the PAF design. Our findings enhance the understanding of FFNs' contributions to the overall performance of transformer models and open up new avenues for future research on improving and optimizing these architectures.


\bibliography{references}

\end{document}